\documentclass[journal, 10pt, times, compsoc]{IEEEtran}
\usepackage{graphicx}
\usepackage{url}
\usepackage{float}
\usepackage{cite}

\ifCLASSOPTIONcompsoc
  \usepackage{cite}
\else
\fi

\ifCLASSINFOpdf

\else
 
\fi

\begin{document}

\title{Behavioural Curves Analysis Using Near-Infrared-Iris Image Sequences}

\author{Leonardo Causa,
        Juan E. Tapia,~\IEEEmembership{Member,IEEE,}
        Enrique Lopez Droguett,
        Andres Valenzuela,
        Daniel Benalcazar,
        and~Christoph Busch,~\IEEEmembership{Senior Member,~IEEE}% <-this % stops a space
\IEEEcompsocitemizethanks{\IEEEcompsocthanksitem Corresponding author: Juan Tapia** and Christoph Busch are with da/sec-Biometrics and Internet Security Research Group, Hochschule Darmstadt, Germany.\protect\\
E-mail: juan.tapia-farias, christoph.busch @{h-da.de}
\IEEEcompsocthanksitem Leonardo Causa, Andres valenzuela and Daniel Benalcazar are with TOC Biometrics, Research and Development Centre, Santiago, Chile.
E-mail: leonardo.causa, andres.valenzuela, daniel.benalcazar @{tocbiometrics.com}
\IEEEcompsocthanksitem Enrique Lopez Droguett is with the Department of Civil and Environmental Engineering, and Garrick Institute for the Risk Sciences, University of California, Los Angeles, USA.}

{\textbf{ }}

{\textbf{This work has been submitted to the IEEE for possible publication. Copyright may be transferred without notice, after which this version may no longer be accessible.}}
% <-this % stops an unwanted space
\thanks{Manuscript received February 2, 2022}}

%{Shell \MakeLowercase{\textit{et al.}}: Bare Demo of IEEEtran.cls for Computer Society Journals}

\IEEEtitleabstractindextext{%
\begin{abstract}
This paper proposes a new method to estimate behavioural curves from a stream of Near-Infra-Red (NIR) iris video frames. This method can be used in a Fitness For Duty system (FFD). The research focuses on determining the effect of external factors such as alcohol, drugs, and sleepiness on the Central Nervous System (CNS). The aim is to analyse how this behaviour is represented on iris and pupil movements and if it is possible to capture these changes with a standard NIR camera. 
The behaviour analysis showed essential differences in pupil and iris behaviour to classify the workers in "Fit" or "Unfit" conditions. The best results can distinguish subjects robustly under alcohol, drug consumption, and sleep conditions. The Multi-Layer-Perceptron and Gradient Boosted Machine reached the best results in all groups with an overall accuracy for Fit and Unfit classes of 74.0\% and 75.5\%, respectively. These results open a new application for iris capture devices.
\end{abstract}

}

\maketitle

\IEEEdisplaynontitleabstractindextext

\IEEEpeerreviewmaketitle

\IEEEraisesectionheading{\section{Introduction}\label{sect:intro}}

\IEEEPARstart{I}{n the} 24/7 society, many people work day and night for public safety and health services or economic reasons. An estimated 15\%-25\% of the workforce works in shifts \cite{BALASUBRAMANIAN202052,ijerph16173081,WICKWIRE20171156}. Working in rotating shifts at night can be a significant risk factor. Also, several studies performed in different regions such as the U.S., Australia, U.K., Japan and others, show an essential correlation between shift-work at night with alcohol and drug consumption, which can be considered as important triggers for occupational accidents \cite{wickwire2017shift}. Workplace alcohol and drug use and impairment directly affect an estimated 15\% of the U.S. workforce; about 10.9\% work under the influence of alcohol or with a hangover \cite{Frone}. The Australian Government alcohol guidelines report shows 13\% of shift-workers, and 10\% of those on standard schedules reported consuming alcohol at risky levels for short-term harm \cite{DorrianandSkinner}. It is essential to point out that these numbers are similar in different parts of the world.

These impairments of fitness for work cause lower performance and an increased likelihood of accidents and are also associated with productivity loss and increased economic costs \cite{ijerph18031294}. Alcohol and substance abuse disorders and their consequences on the heart, liver, immune system, and other organs are on the rise worldwide \cite{Pinheiro2015alcoholvideo}. Which are made worse by the economic and health crisis generated by the COVID-19. According to the National Institute of Drug Abuse (NIDA) of the National Institute of Health’s (NIH), alcohol, illicit drugs and tobacco cost between \$600 billion and \$740 billion per year via lost work productivity, crime, and healthcare \cite{XU2015326, SmokingCessation, sacks2015, NDIC2011, Birnbaum, CurtisF, NIDA2021}. Fatigue is estimated to cost employers another 136 billion dollars \cite{Rosekind, NSC2017, XU2021, Yung2016}. Only in the U.S. alone, the total cost of workplace injuries in 2019 was \$171 billion. The cost per worker was \$1,100, the cost per medically consulted injury was \$42,000, while the cost per death was \$1,2 million. %In the same year, the cost per death was about \$1.2 million. %An estimated total of 105 million workdays were lost in 2019 \cite{NSCInjury2019}. 
Fatigue, drowsiness, and sleepiness caused about 22\% of all injury crashes according to Road Safety Annual Report 2020\footnote{\url{https://www.itf-oecd.org/road-safety-annual-report-2020}} published by the Organisation for Economic Cooperation and Development (OECD). The percentage of fatal crashes caused by fatigue can reach up to 30\%. Reported percentages can differs because it can be difficult to figure out whether fatigue played role in an accident. Nevertheless, these numbers can be underestimated. The World Health Organisation (WHO) reported that fatigue accidents cost most countries 3\% of their gross domestic product \cite{9031734}. Hence, it is necessary to take proactive measures to prevent fatigue, alcohol consumption, and drug abuse from negatively affecting employees and companies.

Currently, many solutions exist for addressing workplace fatigue, alcohol and substance abuse; these solutions are known as Fitness For Duty (FFD) analysis systems \cite{FFD, murphy}. Within the context of occupational testing, FFD describes a toolkit that helps evaluate a person's physical and emotional states by the specific requirements of a job. Being "Fit" means being able to perform the job's duties in a safe, secure, productive, and effective manner. However, these solutions do not adequately addressing all the required elements of workplace accident prevention. 

In order to address these problems, it is necessary to observe what repetitive behaviours or biometric factors manifest through the body to establish the relationship between the cause and effect in a person's behaviour. The Central Nervous System (CNS) controls the iris and pupil movements \cite{Adler1985}. In this condition, the subject can not voluntarily change the pupil's or iris' movement. This action is initiated automatically in response to an external factor such as light, or an internal factor such as alcohol consumption, drug abuse or fatigue. On the other hand, iris recognition allows identifying and distinguishing one worker from another and, thus, guaranteeing that the measurement is from the worker under testing. Therefore, the iris and pupil are highly reliable to measure the fitness for duty of the subject. Hence, there is a need to develop an automated and reliable FFD model tool based on iris recognition framework \cite{Iris-tapia}.

\subsection{Related Work}

Navarro et al. \cite{Navarro7877181} developed a system that captures the driver iris image to detect if the person is drunk. That approach comprises a hardware and software system that implements an algorithm based on the Gabor Filter. The system consists of a Charge-Coupled Device (CCD) Camera and Analog-to-Digital converter linked into a program to process the captured image. The system provides a signal to interact with the car ignition if the software detects that the driver is under the influence of alcohol.

Monteiro et al. \cite{MonteiroPinheiro2015}[25] proposed a non-invasive and simple analysis to detect alcohol use through pupillary reflex analysis. Results presented rates near 85\% correct detections using algorithms for pattern recognition, thus demonstrating the efficacy of the method. The main limitation of this work is related to the active participation of the volunteers, as each subject had to stay in a dark testing room for approximately 5 minutes up to adapt the pupil dilation/constriction to the darkness.

Amodio et al. \cite{Amodio8515064} studied the feasibility of designing a driver alcohol detection system based on the dynamic analysis of a subject's Pupillary Light Reflex (PLR). The test method consists in applying a light stimulus to one eye and to capture the dynamics of constriction of both eyes. For extracting the pupil size profiles from the video sequences, a two-step methodology is described. In the first phase, the iris/pupil search within the image is performed. In the second stage, the image is cropped to perform pupil detection on a smaller image to improve time efficiency. The undesired pupil dynamics arising in the PLR are defined and evaluated. A spontaneous oscillation of the pupil diameter is observed in the range [0, 2] Hz, and the accommodation reflex causes pupil constriction of about 10\% of the iris diameter. 
A set of features is introduced to compare the two populations of responses and is used to design a Support Vector Machine classifier to discriminate between "Sober" and "Drunk" states.

Guddhur et al. \cite{GUDDHURJAYADEV2021} proposed an effective method for the classification of alcohol effect as well as the identification of iris damaged levels utilising a Modified Deep Learning Neural Network (MDLNN). 
Initially, the alcohol image extracts many features such as frequency, shape, and texture. Next, the extracted features are selected utilising Bacterial Foraging Optimisation (BFO). Finally, the segmented iris percentage is estimated based on the Euclidean distance between the original iris image of person under alcohol and the segmented level of the iris image. The proposed work attains better accuracy in an experimental assessment than the prevailing methodologies.

The Psychomotor Vigilance Task (PVT) is a brief vigilance and attention task, and it is considered the gold standard instrument for the assessment of the effects of fatigue \cite{BalkinComparative}. During each 10 minute trial, subjects must attend closely to a stimulus window and respond by pressing a response button. Subjects are instructed to respond as quickly as possible. PVT scores of interest include mean reciprocal reaction time of the slowest 10\% of responses and lapses that correspond to stimulus presentations taking longer than 500 ms. 

In this context, Benderoth et al. \cite{Benderoth3-min} investigated the reliability and validity of a 3-min PVT administered on a portable handheld device, assessing sensitivity to sleep loss and alcohol-related a 10-min PVT and applied tasks. The 3-min PVT showed high reliability and validity in determining sleep loss and alcohol-induced impairments in cognitive performance.

Persson et al. \cite{9055218} studied the reliability of Heart Rate Variability (HRV) as a feature for driver sleepiness detection. Data from real-road driving studies were used, including 86 drivers in both alert and sleep-deprived conditions. Based on the Karolinska Sleepiness Scale (KSS), subjective ratings were used as ground truth. K-nearest neighbours, Support Vector Machine (SVM), AdaBoost, and Random Forest were applied for training and testing the models. The best performance was obtained with the Random Forest classifier with an accuracy of 85\%. The worst results were obtained in the total deprivation sleep group. The results showed that subject-independent sleepiness classification based on HRV performs poorly in realistic driving conditions.

Kim et al. \cite{Kim9427252} studied operator performance in a nuclear power plant using an FFD system using Electroencephalogram (EEG) with a deep learning algorithm to classify an operator's condition. To determine the suitability of this approach, EEG data were collected during simple cognitive exercises designed to examine the mental readiness of nuclear operators. The designed EEG-based FFD classification system could successfully determine an operator's sobriety, stress, and fatigue in a timely and cost-effective manner. This study also investigated schemes for providing information security to the EEG-based FFD status classification system. 

Tanveer et al. \cite{8846024} proposed a deep-learning-based driver-drowsiness detection for brain-computer interface (BCI) using functional Near-InfraRed Spectroscopy (fNIRS). The brain signals were acquired from healthy subjects while driving a car simulator. A Convolutional Neural Network (CNN) was used to classify different alertness conditions. This algorithm was used on colour map images to determine the best suitable channels for brain activity detection in other time windows. The CNN architecture resulted in an average accuracy of 99.3\%, showing that the model was able to differentiate the images of drowsy/non-drowsy states.

Guede-Fernandez et al. \cite{8744224} proposed a drowsiness detection method based on changes in the respiratory signal obtained using an inductive plethysmography belt. The algorithm is based on the analysis of the respiratory rate variability (RRV). Recordings were acquired with a driving simulator cabin and a group of experts rated the drivers' condition of alertness to evaluate the algorithm performance. Results showed a specificity of 96.6\%, a sensitivity of 90.3\%, and Cohen's Kappa agreement score of 0.75 on average across all subjects through a leave-one-subject-out cross-validation.

\subsection{Commercial Devices}

As was mentioned before, various FFD algorithms have been proposed in the literature. Several of these algorithms and concepts were transformed to a product and device to measure the FFD. In this section, we analyse some of these products highlighting the pros and cons of each one.
Some relevant devices are: PMI FIT2000, Sobereye, Optalert \cite{ADA522106}.
The Alertplus, Alermeter \cite{31813584, ADA522106}. Jawbone, Fitbit \cite{Orellana6944547, BaiECG2020, Imboden844, ZambottiM2015, KANDERA2019278}.The Smartcap \cite{Caldwell2009, ButlerP15, Abbood6930193}, among others. 

The PMI-FIT2000\footnote{\url{http://www.pmifit.com/}} uses eye-tracking and pupilometry to identify impaired physiological states due to fatigue and other factors, such as alcohol or drug use. The test requires one minute to complete. The system employs an algorithm that compares an individual’s established baseline to present state on four variables (i.e., pupil diameter, pupil constriction amplitude, pupil constriction latency \& saccadic velocity). The baseline is established by the average of 10 trials taken during non-impaired conditions. After the baseline trials, each subsequent trial provides the user with scores on the four test components plus a composite score, the FIT Index. The PMI-FIT2000 has been used in multiple fatigues and impairment studies in other contexts, such as motor vehicle operation. Note that this system does not perform biometric recognition.

Sobereye\footnote{\url{https://www.sober-eye.com/}} is a portable device used to predict impairment caused by substance abuse or fatigue. It uses a smartphone attached to an opaque enclosure that fits a user’s eyes to measure the PLR. The PLR is an involuntary reflex that changes the size of the pupil when the eyes are exposed to light. A high-intensity light will cause the pupil to constrict, and low-intensity light will cause the pupil to dilate. The user holds the enclosure over the eyes for one minute before any measurements are taken. This time allows the pupils to dilate. After the camera flash turns on for four seconds, a video is taken at 60 frames per second in full high definition resolution ($1920\times1080$pix). Like other FFD examinations identifying a PLR alteration requires establishing a PLR baseline for each employee. The process of establishing a PLR baseline takes 10 days (starts of the workday) to monitor the typical day to day PLR variations. After about ten tests, an employee’s PLR baseline is established. The baseline is then stored and used as a reference for future tests. %In addition, the Initial Pupil Diameter (IPD), Constriction Amplitude (CA), Constriction Velocity (CV), and Latency (L), the Average Value (AVG), and the Standard Deviation (STD) are calculated after each test.
By comparing day to day measurements and calculations to the baseline, PLR alterations can be identified. A standard scoring system is used to evaluate the degree of PLR alteration. An employee can be classified as either “High Risk” or “Low Risk.” This measure indicates the probability of the employee being affected by impairment due to an altered PLR. Additionally, Sobereye uses iris recognition \cite{russo1999saccadic}.

Optalert\footnote{\url{https://www.optalert.com/}} is an Infrared Reflectance (IR) oculography based on the principle that while people are drowsy, the muscle groups controlling eye and eyelid movements are inhibited by the central nervous system \cite{johns2021effects}. IR transducers fitted inside spectacle frames are positioned towards the eye to measure the relative velocity of the opening and closing of the eyelid and blink duration times. A combination of oculometric variables is used to calculate a driver's level of drowsiness in real-time, providing a minute to minute Johns Drowsiness Scale (JDS) rating \cite{JDS2007}. The commercially available system is designed to emit auditory warnings when drivers reach a JDS score of 4.5–4.9 (cautionary level of drowsiness) and a score of 5.0 or above (critical level of sleepiness), associated with an increased risk of severe lane excursions on a driving simulator. The system needs a 5-minute baseline recording before starting each register \cite{FtouniOptalert12}.

AlertMeter\footnote{\url{https://www.deltasleep.ie/alertness-testing/}} is a graphical cognitive alertness test lasting 60-90 seconds. The test interface displays different shapes that the user can identify accurately and quickly. The system does not simulate any particular job function but challenges several key brain functions necessary for all jobs, measuring reaction time, decision-making speed, orientation, and hand-eye coordination. To establish an initial baseline score or individual performance standard, users take the alertness test ten times. The scoring algorithm compares users’ daily test results with their baseline scores. The system identifies compromised alertness when an employee’s test result significantly deviates from their baseline. %Psychological and physiological factors differ greatly between employees, so the only way to accurately measure individuals’ cognitive states is to compare their performance in real-time against their baselines. 
A calculated baseline methodology provides individual feedback rather than a score against an imposed standard. AlertMeter test scores have also been shown to correlate to the day-time, indicating sensitivity to circadian cycles \cite{ferguson2020testing}.

Wearable technologies also allow measurement of clinically-relevant parameters describing an individual's health state. Their varied applications have provided the driving force for the development of a broad range of wearable technologies that can be adapted for use in healthcare, workplace and other fields \cite{8470151}. Wrist wearables such as Jawbone, Apple-Watch, Fitbit and others use sensors that measure heart-rate and motion \cite{iad2021heath, kaewkannate2016comparison, kaewkannate2015review}. Both use Optical Heart Rate monitoring (OHR), or photoplethysmography. The OHR monitor detects pulse by shining a light through the skin to look at blood flow. An accelerometer measures motion by translating movement data into digital measurements. The data from wrist wearables can be used to monitor fatigue in the workplace. Long-standing sleep deprivation is correlated with an increased heart rate, and poor sleep quality leads to a high risk of fatigue.

SmartCap’s LifeBand\footnote{\url{http://www.smartcaptech.com/life-smart-cap/}} is a wearable headband that measures brain EEG. The sensors in the life-band send out filtering (low pass) signals to block signals above 40Hz. Samples at 1280Hz are taken and converted to 256Hz to minimise high-frequency noise. Then, a frequency spectrum of the 256Hz signals is calculated over a five-second time frame. 
As a result, the well-known delta, theta, and alpha waves can be recorded and then scaled by the power of the beta waves. This wave creates a ratio of an individual’s drowsiness and wakefulness. A fatigue score is computed from this ratio, and the risk of fatigue is reported to the worker. This score is calculated based on independently validated algorithms researched by SmartCap \cite{gruenhagen2021technology}.

The study of state of the art in commercial devices of FFD shows several systems based on different technologies. While it is true that they are widely used tools in the industry, their implementation and use have significant limitations:
\begin{enumerate}
    \item Some of these devices can only identify one cause of impairment per time.
    \item The systems use reactive methods, i.e., act once the risk event has been triggered.
    \item Do not report the metrics for the implemented algorithms or their associated scientific validations.
    \item Require the active involvement of workers for the test course. For example, eyes should be closed voluntarily for one minute, which can generate problems in the test results if the involvement is not made in the correct form.
    \item Invasive protocol for capture requires contact between the capture subject and the device.
    \item It is necessary to establish a baseline as reference per subject. Then, the reference is used to compare the scores and evaluate. This situation is a problem because the reference could be intentionally altered.
    \item Most of the systems cannot identify and verify the capture subject performing the test and, therefore, it is possible to impersonate the results using a third person.
\end{enumerate}

A new method based on behavioural curves from NIR iris images is proposed to mitigate the limitations described above. This method allows integrating a FFD system in a unique, light, portable and touch-less mobile device to estimate subjects' proactive alertness before starting their duties and verifying their identity.

It is essential to highlight that our approach will not perform a traditional analysis based on the measurement of alcohol or drug levels in blood using alcohol-test, drug-test or other devices. This research aims to determine the effect of external factors such as alcohol, drugs, and fatigue on the CNS and how this effect is reflected in behaviour changes on pupils and irises diameters and movements.

This paper has four main contributions as follows: 
\begin{enumerate}
    \item  Determine behavioural patterns of the subjects to estimate the trend curves associated with the changes over time in pupil and iris diameters due to the effect of external agents on the CNS.
    \item Develop an algorithm based on iris recognition and pupil behaviour using Near-Infra-Red (NIR) periocular image sequences that determine if the captured subject is in a CNS disturbed condition due to alcohol consumption, drug abuse or sleep deprivation.
    \item  Integrate the behaviour analysis and the detection algorithm in a new FFD model to estimate whether a person is fit or unfit to perform his tasks or duties, without the need for active participation in the subject, without baseline estimation. Moreover, the system is portable and will prevent any impersonation.
    \item Generate a significant annotated database of NIR periocular images including control (normal subjects), alcohol consumption, drug abuse and fatigue conditions to understand the significance of these variations.
\end{enumerate}

The remaining of this article is organised as follows: Section \ref{sec:database} explains the database. The proposed method for FFD detection is presented in Section \ref{sec:method}. The experimental results are discussed in Section \ref{sec:result}. Conclusions and remarks are given in in Section \ref{sec:conclusion}.
\vspace{-0.3cm}

\section{Database}
\label{sec:database}

\begin{figure}[htbp]
\centerline
{\includegraphics[scale=0.41]{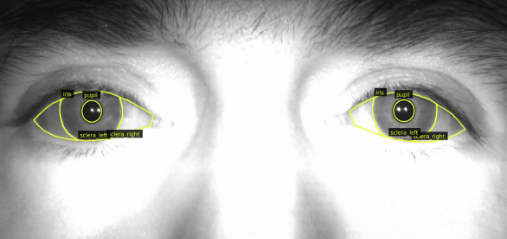}}
\caption{Example of a labelled periocular NIR image. The image shows both eyes and the corresponding labels to the right and left sclera, pupil and iris.}
\label{capture}
\end{figure}

One of the main challenges of this research was to develop a new database. This database contains images of alcohol, drug, and sleepiness information. 
This task was a very demanding effort because all the participants were volunteers. The recruitment process was exhaustive and challenging. We developed a new database called the "FFD NIR iris images Sequences database" (FFD-NIR-Seq) for this research, which contains a set of 10-seconds streams sequences of periocular NIR images. The protocol was analysed and approved by the ethical committee of the University of Chile.
The binocular NIR image sequences captured correspond to the periocular area (eye mask) to obtain both eyes, pupils, iris, and sclera (see Figure~\ref{capture}).

\begin{figure}[H]
\centerline
{\includegraphics[scale=0.34]{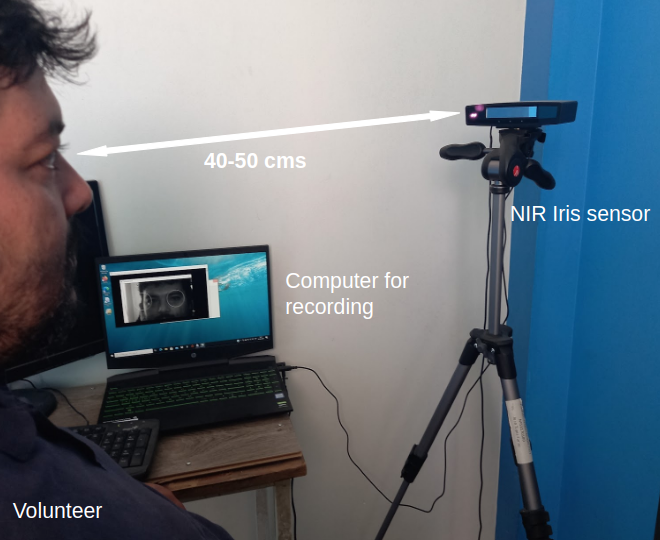}}
\caption{Description of the capture device. The subject is positioned in front of the sensor (40 up to 50 cm). The capture device detects the eyes and starts the recording for 10-sec.}
\label{volunteer}
\end{figure}

Other images were acquired using a monocular capture device to get the area corresponding to each eye separately. For the acquisition of the image sequences, the subject stays in front of the capture device, the equipment detects the eyes and starts the recording (see Figure~\ref{volunteer}).

This data makes it possible to perform the necessary processing to determine the iris and pupil behavioural parameters to be used in developing the models.

\begin{figure}[htbp]
\centerline
{\includegraphics[scale=0.23]{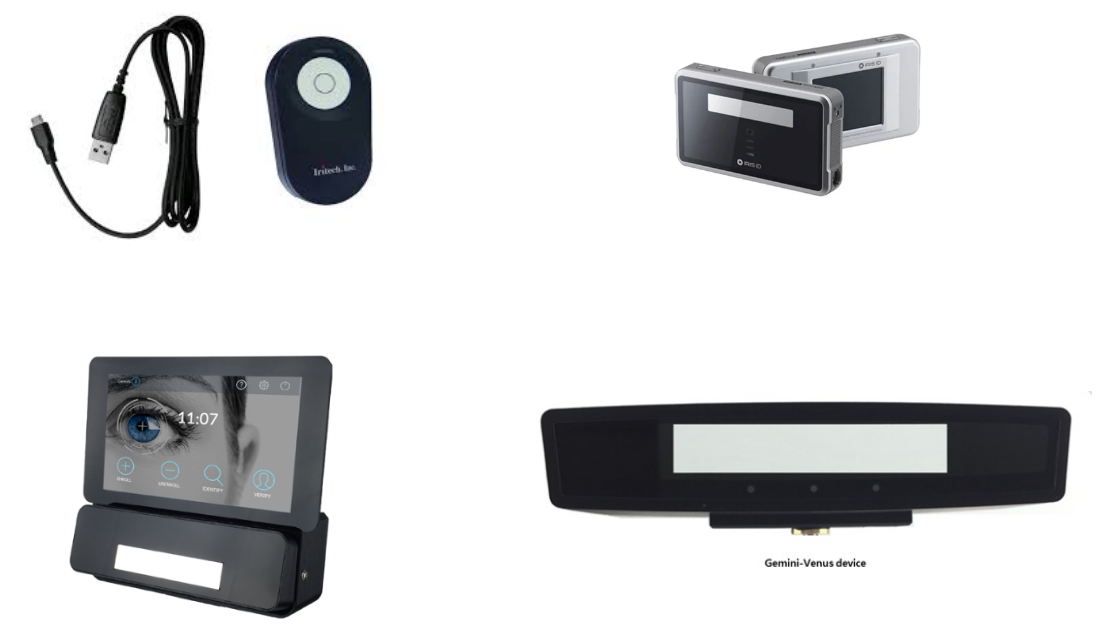}}
\caption{Capture devices used for the acquisition of NIR image sequences. a) Monocular capture device to record each eye area. Capturing devices b)-d) allow for capturing the periocular area (both eyes).}
\label{sensors}
\end{figure}

The images sequences were captured by using four different devices\footnote{\url{https://www.iritech.com}}: i) Iritech MK2120UL (monocular), ii) iCAM TD-100A, iii) Iritech Gemini, and iv) Iritech Gemini-Venus. Figure~\ref{sensors} shows the NIR capture devices used for database acquisition. The room temperature and lighting (200 lux) were kept constant in the capturing process.

Four NIR image sequences in different conditions were registered:
\begin{itemize}
   \item 
Control DB: healthy subjects that are not under alcohol and/or drugs influence and in normal sleeping conditions.
   \item 
Alcohol DB: subjects who have consumed alcohol or are in an inebriation state.
   \item 
Drugs DB: subjects who have consumed some drugs (mainly marijuana) or who consume psychotropic drugs (by medical prescription).    
   \item 
Sleep DB: subjects with sleep deprivation, resulting in fatigue and/or drowsiness due to sleep disorders related occupational factors (shift structures with high turnover).
\end{itemize}

\subsubsection{Alcohol}
In the case of the alcohol database, the subjects were submitted to the following protocol: 
\begin{enumerate}
    \item The first NIR image sequence acquisition was made at time 0 (previous to alcohol consumption).
    \item All the Volunteers drank 200 ml of alcohol in up to 15 minutes. 
    \item The second acquisition was performed immediately after the alcohol intake was finished, i.e., 15 minutes after 0.
    \item The third acquisition was made 30 minutes after time 0.
    \item Fourth acquisition was made 45 minutes after time 0. 
    \item Finally, the fifth acquisition was made 60 minutes after time 0. 
\end{enumerate}
Thus, there were five sequences recorded of images of the subject under the effects of alcohol and one sequence of control images.

\subsubsection{Drugs}
According to the World Drug Report, 2021, of the United Nations Office on Drugs and Crime, cannabis\footnote{\url{https://www.unodc.org/unodc/en/data-and-analysis/wdr2021.html}} is the most widely consumed crop worldwide with an annual prevalence of 15\%, followed by pharmaceutical opioids and tranquillisers with a 5\% and 2.5\% of yearly prevalence, respectively. For this reason, about 95\% of the records in our database correspond to cannabis consumption. In contrast, the remaining 5\% corresponds to tranquillisers and more complex drugs (heroin and ecstasy). 
The volunteers were drug consumers for the drug database acquisitions, and the images recordings take place at least 30 minutes after the initial consumption. 

\subsubsection{Sleep}
For the sleep database, a particular image acquisition protocol was defined, in which tests were performed under controlled sleep deprivation conditions. These recordings were obtained on a specific group of subjects who were subjected to different levels of sleep deprivation to evaluate the level of fatigue/drowsiness at different time intervals. The volunteers were monitored by using a smart band to measure the quantity and quality of sleep.
Subjects were grouped as follows:
\begin{enumerate}
    \item Total sleep deprivation
    \item Less than 3 hours of night sleep
    \item Between 3 and 6 hours of night sleep
    \item More than 6 hours of sleep (normal sleep)
\end{enumerate}

%The following variables were recorded with the smart band:

%\begin{enumerate}
%    \item Hours of sleep
%    \item Quality of sleep
%    \item Quantity of intra-sleep wakefulness
%    \item Duration of intra-sleep wakefulness
%    \item Naps
%    \item Hypnogram (evolution of sleep states and stages)
%\end{enumerate}

\begin{figure}[htbp]
\centerline
{\includegraphics[scale=0.40]{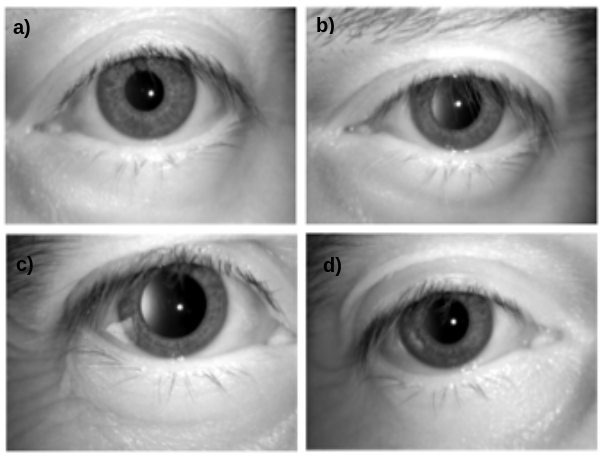}}
\caption{Examples of the NIR images captured. a) Control, b) Alcohol, c) Drug, and d) Sleep images.}
\label{img_examples}
\end{figure}

During the recording season, volunteers performed three daily image acquisitions: i) beginning of the working day, ii) post-lunch, and iii) at the end of the working day.
Figure~\ref{img_examples} shows examples of the acquired images for each of the conditions defined in the database.

The FFD-NIR-Seq database consists of 1,510 eye-disjoint images (with 144,011 individual images). On average, 150 images are captured per subject. This process took ten seconds. The image sequences were divided into three sets: training, validation and test. In the case of the test set, the aim was to represent the actual proportion of unfit instances, close to 15\%. Table~\ref{tab1} shows the number of sequences, and Table~\ref{tab2} shows the total images, in both cases, organised by conditions.

\begin{table}[H]
\centering
\caption{NIR sequences by condition.}
\begin{tabular}{l|c|c|c}
\hline
\textbf{Conditions} & \textbf{Train set} & \textbf{Validation set} & \textbf{Testing set} \\ \hline
\hline
\textbf{Control} & 247 & 35 & 688 \\ \hline
\textbf{Alcohol} & 247 & 35 & 72 \\ \hline
\textbf{Drug} & 62 & 9 & 17 \\ \hline
\textbf{Sleep} & 69 & 9 & 20 \\ \hline
\textbf{Total} & 625 & 88 & 797 \\ \hline
\end{tabular}
\label{tab1}
\end{table}

\begin{table}[htbp]
\centering
\caption{Description of total NIR images recording by condition.}
\begin{tabular}{l|c|c|c}
\hline
\textbf{Conditions} & \textbf{Train set} & \textbf{Validation set} & \textbf{Testing set} \\ \hline
\hline
\textbf{Control} & 21,449 & 3,136 & 60,222 \\ \hline
\textbf{Alcohol} & 24,325 & 3,394 & 6,998 \\ \hline
\textbf{Drug} & 8,653 & 1,253 & 2,338 \\ \hline
\textbf{Sleep} & 8,568 & 1,140 & 2,535 \\ \hline
\textbf{Total} & 62,995 & 8,923 & 72,093 \\ \hline
\end{tabular}
\label{tab2}
\end{table}

This database was used to train and validate the different stages of the FFD model, including the object detection, segmentation, and final classification stages.

\section{Methodology}
\label{sec:method}

\begin{figure}[H]
\centerline
{\includegraphics[scale=0.40]{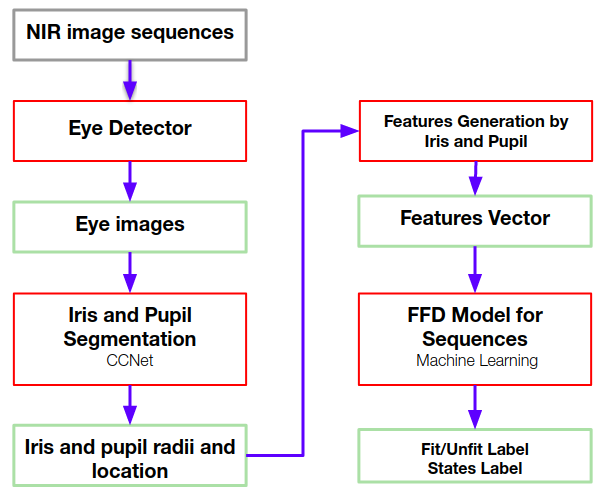}}
\caption{Block diagram of the proposed FFD model based on images sequences. The input is NIR image sequences followed by the eyes detector, eye segmenter, feature extraction, and the FFD model. The output vector is the input to the machine learning classifier. The system delivers the FFD level: the fit/unfit indicator.}
\label{method}
\end{figure}

The proposed FFD model can be described as an analysis cascade of four modules, as shown in Figure~\ref{method}. The eye detector module allows one to find the eyes and crop them to the right and left instance (in periocular images). The iris and pupil segmentation module is applied to the single eye image sequences to generate the iris and pupil variables: radii and centres are estimated. 
A feature extraction module uses the iris and pupil measures to create a feature vector that represent the temporal behaviour of pupil and iris radii over the acquired image sequences. Finally, the FFD model generates the final classification: Fit/Unfit indicator and level for each state (control, alcohol, drug and sleep).

\subsection{Eye Detector}

The eye detector module was implemented to find both eyes in the input periocular images and for subsequent cropping and segmentation. This algorithm is detailed in our previous work on semantic segmentation of periocular NIR images under alcohol effects \cite{tapia2021semantic}. This module applied Eye-tiny-yolo, classical tracking, and semantic segmentation by Cluster-Coordinated Net \cite{ccnet} (CCNet).

\begin{figure}[htbp]
\centerline
{\includegraphics[scale=0.32]{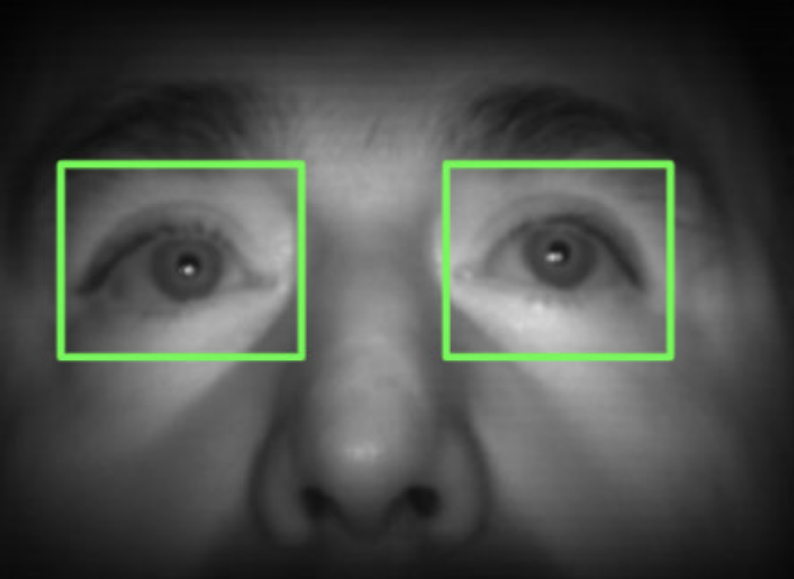}}
\caption{Example of Eye-tiny-yolo detection module applied on periocular images with both eyes automatically detected.}
\label{eye_detector}
\end{figure}

First, Eye-tiny-yolo detects the left and right eyes represented as a rectangular areas. Then, a Multiple Instance Learning (MIL) \cite{maron1998framework} and Channel and Spatial Reliability Tracker (CSRT) \cite{farkhodov2020object} tracking methods were applied to detect interest points and descriptors to track each eye frame by frame. Finally, we use a modified version of CCNet to define the final rectangular area that contains each eye and crop the images again (see Figure~\ref{eye_detector}).

\subsection{Iris and Pupil Segmentation}

\begin{figure}[htbp]
\centerline
{\includegraphics[scale=0.45]{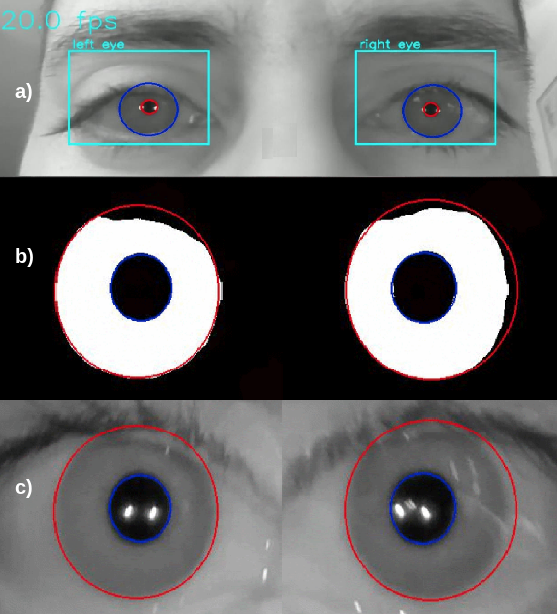}}
\caption{Example of iris and pupil segmentation process. a) Periocular NIR image with the Eye-tiny-yolo detector, b) Mask for pupil and iris, and c) Result of the segmentation process to determine the iris and pupil circles used to define the radii and centres.}
\label{segment}
\end{figure}

Once both eyes are cropped (left and right), we applied a semantic segmentation method trained from scratch called CCNet to find a mask that segments the iris and the pupil. The CNN was trained using a subset of monocular NIR images with an aggressive data augmentation process with non-geometrical transformations. These networks output a mask that highlights pixels that belong to the iris and the pupil in the images. In order to use this information, we employ pupil and iris localisation algorithms, which find the centres and the radii of the circles that best adapt the pupil and iris contours by using the mask as input. This method employs the morphological erosion operation and a $XOR$ function to find the shape of the valid iris area. Then, the mean square error was used to determine the circle that best fits the right pupil and iris. The outputs are the radii and centres of pupil and iris in pixels for each image in the sequences (see Figure~\ref{segment}).

\subsection{Feature Generation by Iris and Pupil}

\begin{figure}[htbp]
\centerline
{\includegraphics[scale=0.42]{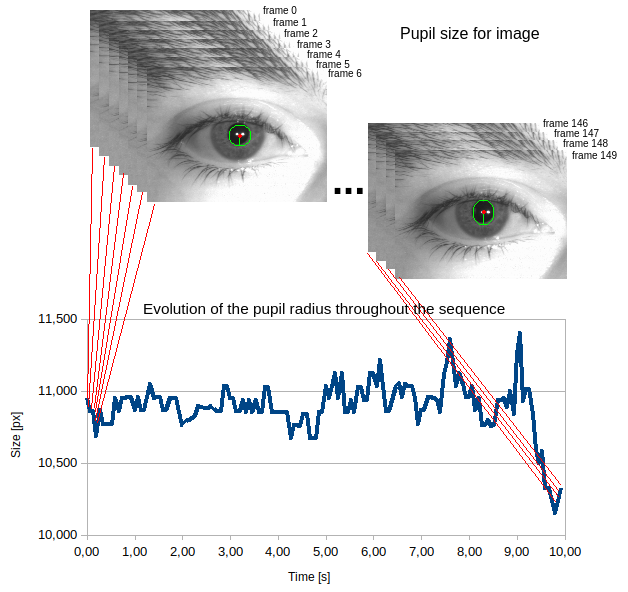}}
\caption{Example of the process to transform individual measurements per image to a time sequence per capture.}
\label{img2seq}
\end{figure}

Once the radii and centres measures for pupil and iris were obtained for all image in each sequence, we estimated the features based on the variation in pupil and iris radii for each time. In this way, the information from the 100 images is transformed into one (or several) associated vectors with the recording and no longer with the image itself. See Figure~\ref{img2seq}.

The results showed significant inter-class differences in the behaviour between the fit (control) and unfit (alcohol, sleep and drug) class. Feature extraction and selection methods were applied to the input vector to the sequence machine learning model. The feature vector comprises 50 variables representing pupil behaviour throughout the captures. These features vector are the input for the classification model. 

The pupil-iris ratio also was used to generate the feature vectors to train and test the machine learning models. The pupil-iris ratio allows us to eliminate pupil distortion in the computation and iris radius due to the subject's distance from the capture device.

The variables of the feature vector is based on all frames captured within one session at the frame-rate of 15 fps and can be described as follows:

\begin{itemize}
   \item \textbf{Sequence trend}: It allows for the evaluation of the temporal behaviour of the sequence of frames (stemming from one single capture session). It is estimated using the least-squares method to determine the slope $m$ and the intercept $b$ of the series:
   \[ m=\frac{n \cdot \sum({x_i} \cdot {y_i})-\sum {x_i} \cdot \sum {y_i}}{n \cdot \sum {x_i}^2 - |\sum {x_i}|^2} \]
   \[ b=\frac{\sum {y_i} \cdot \sum {x_i}^2 - \sum {x_i} \cdot \sum({x_i} \cdot {y_i})}{n \cdot \sum {x_i}^2 - |\sum {x_i}|^2} \]
   
   where, ${x_i}$ represents the time period at instant $i$ (independent variable), ${y_i}$ the value of the function at the corresponding time stamp at the instant $i$ (dependent variable) and $n$ corresponds to the number of observations.
   
   \item \textbf{Starting sequence trend}: It corresponds to the estimation of the trend, but it is measured in the first second of the sequence. In the unfit cases, the device lights produce a slower change in the pupil-iris ratio than in the fit cases, resulting in a steeper slope for the control subjects. It is calculated using the same definition as described above.
   
   \item \textbf{Moving sequence trend}: The linear regression was estimated using a displacing rectangular window of one second with an overlap of 0.5 seconds between windows.
   
   \item \textbf{Distance to the representative curves}: The distance between the analysed sequence and each of the representative curves of the classes is calculated. Both sequence and classes lines are estimated by linear regression.
   This distance $d$ is obtained according to the following equation. See Figure~\ref{distance}.
   \[ d(r,s)=\frac{|[\overrightarrow{v_r}, \overrightarrow{v_s}, \overrightarrow{P_r P_s}]|}{|\overrightarrow{v_r} \times \overrightarrow{v_s}|} \]
   
   where, $r$ and $s$ represent the lines, $\overrightarrow{v_r}$ y $\overrightarrow{v_s}$ correspond to the direction vectors of lines $r$ and $s$ respectively, and $\overrightarrow{P_r P_s}$ corresponds to the vector formed by a point on each line.
   
   \begin{figure}[htbp]
\centerline
{\includegraphics[scale=0.43]{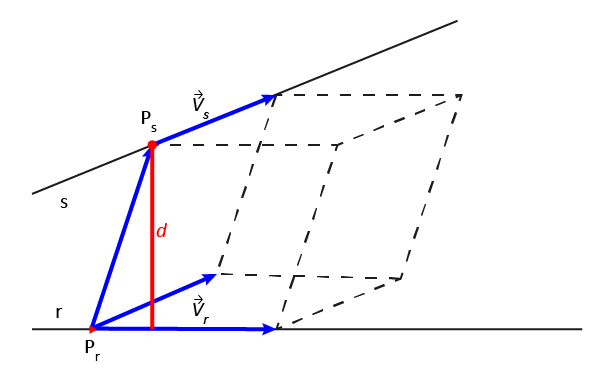}}
\caption{This metric represents the estimation distance between two skew lines. This example defines the equation parameters to compute the distance to the representative curve for each class in the feature vector.}
\label{distance}
\end{figure}
 
   \item \textbf{Statistics values}: Some statistic measures were calculated for each sequence:
        \begin{itemize}
        \item Arithmetic Mean ($\mu_{seq}$): represents the sum of values of the sequence divided by the numbers of elements.
        \item Standard Deviation ($\sigma_{seq}$): represents the quantified variation or dispersion of values in the sequence.
        \item Range: represents the difference between the lowest and highest values in the sequence.
        \item Coefficient of Variation: measurements of the relative data dispersion points in the sequence around the mean:
        \[ CV_{seq}=\frac{\sigma_{seq}}{\mu_{seq}} \]
        \item Coefficient of Variation to the representative curves: measure of the relative dispersion data points in the analysed sequence around the mean of each of the representative curves in each condition:
        \[ CV_{pb_i}=\frac{\sigma_{pb_i}}{\mu_{seq}} \]
        where i = \{control, alcohol, drug, sleep\}
        \end{itemize}
   
   \item Pupil-Iris Ratio values: this metric represents the real-time ratio between the pupil radius and the iris value. It corresponds to each of the values of the first 5 seconds of the sequence.
   
\end{itemize}

\subsection{FFD Model for Image Sequences}

The temporal changes variables related to the iris and pupil radii were used to train and test three different machine learning models. 
The classifier performs the estimation of "Fit" and "Unfit" groups. In addition, these models were also used to classify each of the possible states (control, alcohol, drugs and sleep). The algorithms tested were: Random Forest (RF) \cite{breiman2001random}, Gradient Boosting Machine (GBM) \cite{friedman2001greedy}, and Multi-Layer Perceptron (MLP) Neural Network \cite{rosenblatt1958perceptron}. The models were trained and tested with the sequence sets showed in Table~\ref{tab1}.

RF and GBM were selected because they show promising results for small data sets. This characteristic is essential in the cases of drug and sleepiness recordings, where the pupil size is considerably smaller than in the control and alcohol cases. On the other hand, RF, GBM and MLP are widely tested techniques in unbalanced databases, as in this case, where unfit subjects represent about 13\% of the records, which is consistent with the known statistics about alcohol, drugs consumption and the fatigue presence in work spaces.

Optimal values of model hyper-parameters were obtained using a grid search: an exhaustive search performed on the specific parameter values of a model. Tables~\ref{rf_params}, \ref{gbm_params} and \ref{mlp_params} show the selected hyper-parameters that maximised the precision score.

\begin{table}[htb]
\centering
\caption{Random Forest parameters.}
\label{rf_params}
\begin{tabular}{|lc|}
\hline
\multicolumn{2}{|c|}{\textbf{Random Forest}}                                       \\ \hline
\multicolumn{1}{|c|}{\textbf{Parameters}}                         & \textbf{Value} \\ \hline
\multicolumn{1}{|l|}{Number of estimators (n\_estimators)}         & 1.000          \\ \hline
\multicolumn{1}{|l|}{Criterion (criterion)}                       & "entropy"      \\ \hline
\multicolumn{1}{|l|}{Maximum depth of the tree (max\_depth)}      & 5              \\ \hline
\multicolumn{1}{|l|}{Minimum samples split (min\_samples\_split)} & 5              \\ \hline
\multicolumn{1}{|l|}{Minimum samples leaf (min\_samples\_leaf)}   & 3              \\ \hline
\multicolumn{1}{|l|}{Maximum number of features (max\_features)}  & "auto"         \\ \hline
\multicolumn{1}{|l|}{Cross-validation (KFold)}                    & 5              \\ \hline
\end{tabular}
\end{table}

\begin{table}[htb]
\centering
\caption{Gradient Boosting parameters.}
\label{gbm_params}
\begin{tabular}{|lc|}
\hline
\multicolumn{2}{|c|}{\textbf{Gradient Boosting Classifier}}                                             \\ \hline
\multicolumn{1}{|c|}{\textbf{Parameters}}                                            & \textbf{Value}   \\ \hline
\multicolumn{1}{|l|}{Loss function (loss)}                                           & "deviance"       \\ \hline
\multicolumn{1}{|l|}{Learning rate (learning\_rate)}                                 & 0.01             \\ \hline
\multicolumn{1}{|l|}{The number of boosting stages (n\_estimators)}                  & 1.000            \\ \hline
\multicolumn{1}{|l|}{Subsample (subsample)}                                          & 1.0              \\ \hline
\multicolumn{1}{|l|}{Criterion (criterion)}                                          & "squared\_error" \\ \hline
\multicolumn{1}{|l|}{Minimum samples split (min\_samples\_split)}                    & 10               \\ \hline
\multicolumn{1}{|l|}{Minimum samples leaf (min\_samples\_leaf)}                      & 5                \\ \hline
\multicolumn{1}{|l|}{Maximum depth of estimators (max\_depth)} & 5                \\ \hline
\multicolumn{1}{|l|}{Maximum number of features (max\_features)}                     & "auto"           \\ \hline
\multicolumn{1}{|l|}{Cross-validation (KFold)}                                       & 5                \\ \hline
\end{tabular}
\end{table}

\begin{table}[htb]
\centering
\caption{Neural Network parameters}
\label{mlp_params}
\begin{tabular}{|lc|}
\hline
\multicolumn{2}{|c|}{\textbf{Gradient Boosting Classifier}}                           \\ \hline
\multicolumn{1}{|c|}{\textbf{Parameters}}                            & \textbf{Value} \\ \hline
\multicolumn{1}{|l|}{Hidden layer Neurons (hidden\_layer\_sizes)} & (25, 10)       \\ \hline
\multicolumn{1}{|l|}{Activation function (activation)}               & "relu"         \\ \hline
\multicolumn{1}{|l|}{Solver for weight optimisation (solver)}        & "adam"         \\ \hline
\multicolumn{1}{|l|}{L2 penalty (alpha)}                             & 10e-5          \\ \hline
\multicolumn{1}{|l|}{Size of mini batches (batch\_size)}              & "auto"         \\ \hline
\multicolumn{1}{|l|}{Learning rate (learning\_rate)}                 & "constant"     \\ \hline
\multicolumn{1}{|l|}{Initial learning-rate (learning\_rate\_init)}   & 10e-3          \\ \hline
\multicolumn{1}{|l|}{Maximum number of iterations (max\_iter)}       & 300            \\ \hline
\multicolumn{1}{|l|}{Cross-validation (KFold)}                       & 5              \\ \hline
\end{tabular}
\end{table}

\section{Experimental Results}
\label{sec:result}

\subsection{Behavioural Analysis}
%Figure 10
\begin{figure*}[]
\includegraphics[scale=0.60]{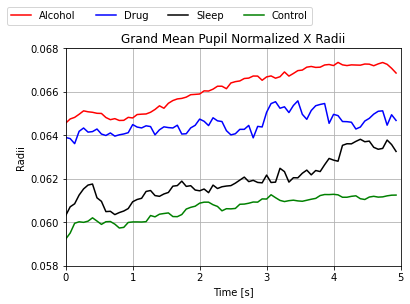}
\includegraphics[scale=0.60]{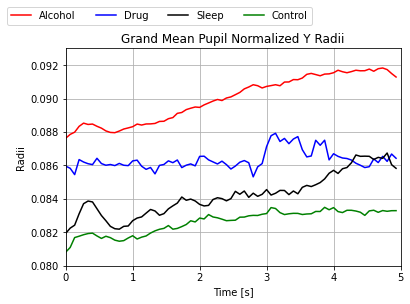}
\caption{Grand mean curves for the normalised \textbf{pupil} radius on the alcohol (red), drug (blue), sleep (black) and control (Green). Left: Horizontal axis (X). Right: Vertical Axis (Y).}
\label{pupil}
\end{figure*}

%Figure 11
\begin{figure*}[]
\includegraphics[scale=0.6]{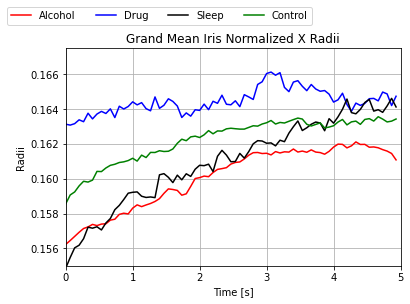}
\includegraphics[scale=0.6]{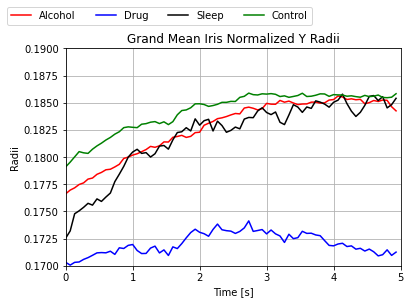}
\caption{Grand mean curves for normalised \textbf{iris} radius on the alcohol (red), drug (blue), sleep (black) and control (Green). Left: Horizontal axis (X). Right: Vertical Axis (Y).}
\label{iris}
\end{figure*}

The study of the behavioural reaction on the training set was performed to determine differences in the size of the pupil and iris radius in conditions of alcohol and drug consumption or sleep deprivation (unfit) versus control subjects (fit). This analysis was performed using the grand mean algorithm. The grand mean of a set of multiple sub-samples is the mean of all observations: every data point, divided by the joint sample size \cite{8954814}.

To obtain the control and alcohol, drug and sleep class curves, the grand mean algorithm was estimated for the pupil radii for each time for all the subjects in the training set in each of the groups. Thus, it is possible to define the baseline behaviour curve for people in Fit and Unfit conditions. 
Note that this analysis shows the average behaviour, so it is not possible to use it as a single variable to separate the classes.

Figure~\ref{pupil} present two plots with the differences in the temporal behaviour of pupil for X and Y axes, with a greater radius size (more significant dilation) in alcohol and drug subjects, followed by sleep deprivation records. In contrast, control subjects have the lowest dilation levels. All curves show an initial period with a pronounced increase in pupil radius size due to the natural adjustment process by the effect of the capture device. Then, control subjects tend to stabilise and produce a slower dilation level. In contrast, in the unfit cases, the dilation trend curves present a higher slope, being especially pronounced in the case of the alcohol records. The behaviour of sleep recordings is closer to control subjects.

\begin{figure*}[]
\includegraphics[scale=0.6]{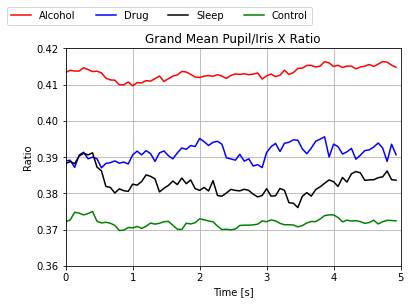}
\includegraphics[scale=0.6]{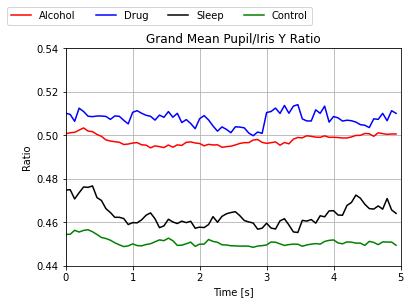}
\caption{Grand mean curves for \textbf{pupil-iris ratio} on the alcohol (red), drug (blue), sleep (black) and control (Green). Left: Horizontal axis (X). Right: Vertical axis (Y).}
\label{pupil_iris}
\end{figure*}

\begin{figure*}[]
\includegraphics[scale=0.6]{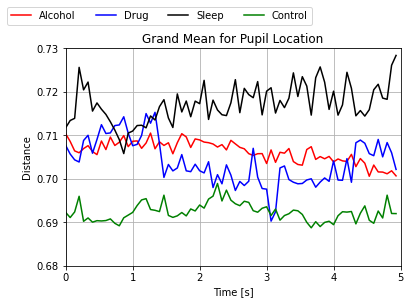}
\includegraphics[scale=0.6]{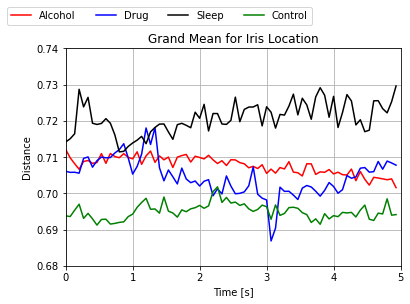}
\caption{Grand mean curves for centring subject on \textbf{horizontal localisation} for pupil and iris on images on the alcohol (red), drug (blue), sleep (black) and control (green) sets. Left: Pupil. Right: Iris.}
\label{pupil_location}
\end{figure*}

Figure~\ref{iris} shows the behaviour of the iris throughout the sequence. In contrast to the pupil (Figure~\ref{pupil}), the iris does not show a significant variation for control subjects with respect to the other classes. This makes sense, as the iris size should not change, except in some pathologies such as uveitis (inflammation of iris). Although the curves show a slight shift in the iris size, this effect should be to the unfit subjects; these tend to keep closed eyes and blink more times, and in many cases, to move in front of the capture device. This action makes the segmentation process difficult and, therefore, the estimation of the iris radius. This effect is not evident in the case of the pupil due to its smaller size. 
In the case of the radii values measured along the horizontal axis, as pointed out above, it is not possible to establish significant differences between the classes, which is correct. Also, there is no eye closure effect in this direction. However, the impact of partial eye closure appears in the case of iris radius estimation along the vertical axis. This effect is more marked in drugged subjects, who present a completely different curve than the other classes, with considerably smaller iris sizes. This effect is because subjects under the influence of drugs (mainly marijuana) tend to have droopy eyelids, and therefore the estimation of the iris size is affected. On the other hand, the difference in the beginning values recording in both analysed axes is due to the natural adjustment process subject in front of the capture device.

The Eye-tiny-yolo model tries to adjust the cropping due to the subject's distance concerning the capture device. In order to mitigate this problem, the ratio between pupil and iris radius was studied as the quotient for each time of both measurements. In this way, the effect of the subject's distance from the camera is eliminated.

For both the estimation on the horizontal and vertical axes, it can be observed that the temporal behaviour of the groups is entirely different, showing a clear separation between them (see Figure~\ref{pupil_iris}).

The control group shows the lowest pupil-iris ratio, which is consistent with the behaviour of pupil radius (Figure~\ref{pupil}). In the case of sleep-deprived subjects, a difference can be observed concerning the control group, but not very marked. This difference could be due to the low number of recordings of this type or the lack of subjects with total sleep deprivation or night shift schedule work. Subjects who have consumed alcohol and drugs show the highest ratios, which means a very significant pupil dilation. The iris size effect described above for the drug recordings on the vertical axis is reflected in the estimation of the pupil-iris ratio in this group, showing the most significant difference when the analysis is made along this axis.

\begin{figure}[htb]
\includegraphics[scale=0.49]{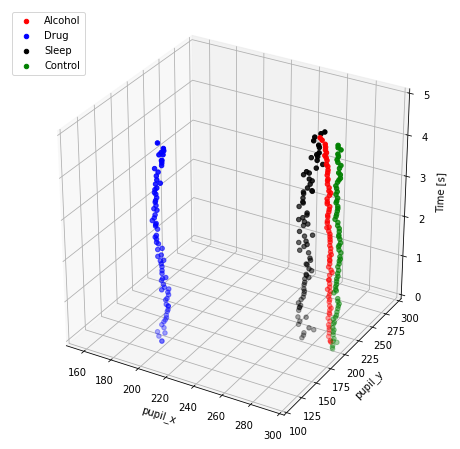}
\caption{Grand mean curves for \textbf{pupil centre position} prior segmentation image on the alcohol (red), drug (blue), sleep (black) and control (Green) sets.}
\label{pupil_mov}
\end{figure}

Figures~\ref{pupil_location} and~\ref{pupil_mov} show a movement analysis of the subjects in front of the captured device at the recorded time. It is expected that fit subjects can keep a better posture in front of the capture device during the recording than unfit subjects who should exhibit more significant movements in all directions. 

Figure~\ref{pupil_location} presents the estimation of the distance of the pupil and iris centres concerning the (0,0) position of the image. It can be seen that in the control cases, the subject tends to maintain a more stable posture throughout the registration. In contrast, the behaviour is more erratic for the unfit subjects, with constant movements in front of the capture device and moving away from the established reference point.

%%%

%Figure~\ref{pupil_location} 
Figure~\ref{pupil_mov} shows a similar analysis. In this case, the position of the pupil centre concerning the image is observed. The control and alcohol subjects show similar and relatively stable behaviours for the pupil centre position in the image. Conversely, the subjects with sleep deprivation and those who have consumed drugs show more eye movements in the image. Moreover, the centre position is completely displaced in the horizontal axis for the drugs case, which shows the subject's movement in front of the capture device. 

The analysed curves show a differentiated behavioural among the groups. Thus, it is possible to use these values as baselines for the conduct of the subjects in the different states. In this way, it is possible to eliminate the process of defining baselines per individual (as commercial solutions do), which can be significantly prolonged and susceptible to being impersonated.

\subsection{FFD Model for Image Sequences: Results}

The system was trained and the parameters were adjusted employing an iterative process using the training and validation datasets. The final performance was measured using the test set.

\subsubsection{Results for each condition}

Tables~\ref{tab_rf_all}, ~\ref{tab_gb_all} and ~\ref{tab_mlp_all} present the statistics results for each model. The metrics used are the following:

\[ Sensitivity=\frac{TP}{TP+FN} \]
\[ Specificity=\frac{TN}{TN+FP} \]
\[ F1-score=\frac{TP}{TP+\frac{1}{2}*(FP+FN)} \]
\[ Accuracy=\frac{TP+TN}{TP+FP+FN+TN} \]

where, $TP$ is a True Positive, $TN$ is a True Negative, $FP$ is a False Positive and $FN$ is the False Negative. 

\begin{table}[htbp]
\centering
\caption{RF statistics results.}
\begin{tabular}{l|c|c|c|c}
\hline
\textbf{Class} & \textbf{Sensitivity} & \textbf{Specificity} & \textbf{F1-score} & \textbf{Accuracy} \\ \hline
\hline
\textbf{Control} & 70.1\% & 94.7\% & 80.5\% & 70.8\% \\ \hline
\textbf{Alcohol} & 73.6\% & 20.5\% & 32.1\% & 71.9\% \\ \hline
\textbf{Drug} & 41.2\% & 53.8\% & 46.7\% & 98.0\% \\ \hline
\textbf{Sleep} & 20.0\% & 23.5\% & 21.6\% & 96.4\% \\ \hline
\end{tabular}
\label{tab_rf_all}
\end{table}

\begin{table}[htbp]
\centering
\caption{GBM statistics results.}
\begin{tabular}{l|c|c|c|c}
\hline
\textbf{Class} & \textbf{Sensitivity} & \textbf{Specificity} & \textbf{F1-score} & \textbf{Accuracy} \\ \hline
\hline
\textbf{Control} & 73.1\% & 95.8\% & 82.9\% & 74.0\% \\ \hline
\textbf{Alcohol} & 76.4\% & 22.9\% & 35.3\% & 74.7\% \\ \hline
\textbf{Drug} & 47.1\% & 53.3\% & 50.0\% & 98.0\% \\ \hline
\textbf{Sleep} & 25.0\% & 29.4\% & 27.0\% & 96.6\% \\ \hline
\end{tabular}
\label{tab_gb_all}
\end{table}

\begin{table}[htbp]
\centering
\caption{MLP neural network statistics results.}
\begin{tabular}{l|c|c|c|c}
\hline
\textbf{Class} & \textbf{Sensitivity} & \textbf{Specificity} & \textbf{F1-score} & \textbf{Accuracy} \\ \hline
\hline
\textbf{Control} & 75.3\% & 95.4\% & 84.2\% & 75.5\% \\ \hline
\textbf{Alcohol} & 70.8\% & 22.9\% & 34.6\% & 75.8\% \\ \hline
\textbf{Drug} & 29.4\% & 29.4\% & 29.4\% & 97.0\% \\ \hline
\textbf{Sleep} & 25.0\% & 35.7\% & 29.4\% & 96.9\% \\ \hline
\end{tabular}
\label{tab_mlp_all}
\end{table}

The general statistics show that the control and alcohol classes obtained the best performances. These results, obtained using a significantly higher number of captured images, belong to these classes. Conversely, drugs and sleep classes present a fewer number of images. 

In the case of sensitivity, the conditions for all trained models show values higher than 70\% in control and alcohol classes. The performance decreased for drugs for which was obtained on average 40.0\%, followed by the sleep class around 25\%. 

In the case of specificity, the results for the control group in all models are close to 95\%, which indicates that the model can correctly detect these subjects, which means that the operation would not be impaired blocking healthy (fit) subjects. 

In the cases of alcohol, drugs, and sleep, the specificity decreases. Although the results are outstanding for the overall accuracy, the test set is unbalanced among the four classes. The database imbalance is consistent with reality since it is estimated that in a real operation, between 10\% to 15\% of the subjects would not be able to perform their tasks correctly due to some external agent such as alcohol, drugs, or sleep.

\subsubsection{Fit and Unfit classes results}

\begin{figure}[]
\centerline
{\includegraphics[scale=0.54]{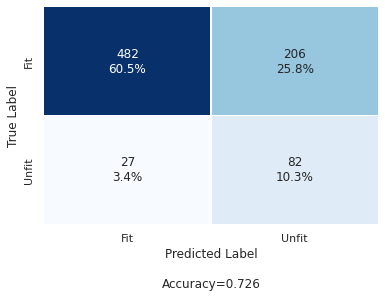}}
\caption{RF confusion matrix for Fit and Unfit classes. Results show the number of cases and the percents in each case.}
\label{rf_afu}
\end{figure}

\begin{figure}[]
\centerline
{\includegraphics[scale=0.54]{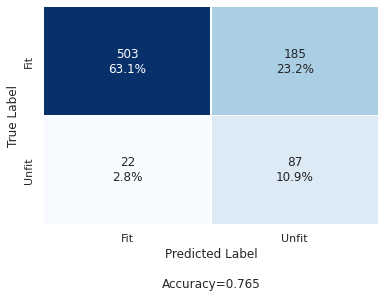}}
\caption{GBM confusion matrix for Fit and Unfit classes. Results show the number of cases and the percents in each case.}
\label{gb_fu}
\end{figure}

\begin{figure}[]
\centerline
{\includegraphics[scale=0.54]{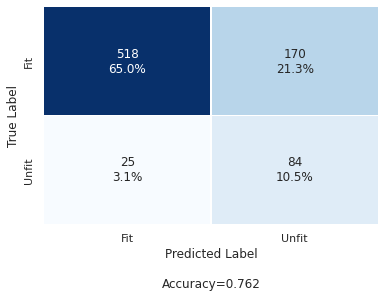}}
\caption{MLP neural network confusion matrix for Fit and Unfit classes. Results show the number of cases and the percents in each case.}
\label{mlp_fu}
\end{figure}

To analyse the results for the Fit and Unfit classes, no new models were trained, but the unfit states were grouped, i.e., the results for alcohol, drugs and sleep are considered as a single class. Figures~\ref{rf_afu}, ~\ref{gb_fu} and ~\ref{mlp_fu} show the classification results using confusion matrices for classes fit and unfit, which present the number of cases per class, the associated percentages and the mean accuracy per class.

As expected, the results improved significantly by grouping the unfit subjects into a single class. For the confusion matrices, the average accuracy per class increases to 72.6\% for RF, 76.5\% for GBM and 76.2\% for MLP. These results increase by almost 25\% concerning the evaluation with four conditions.

Tables~\ref{tab_rf_fu}, ~\ref{tab_gb_fu} and ~\ref{tab_mlp_fu} present the statistics results for each model. In this analysis, all the general statistics show significant improvements. The sensitivity in both classes and for all models exceeds 70\%, reaching 75.3\% in the case of MLP for fit and 77.1\% for the unfit class.

The specificity for the fit class is close to 95\% for all models, and the unfit class ranges between 28.5\% and 40.0\%. 
In the case of overall accuracy, since one is dealing with balanced classes, the results are significant, showing above 70\% in all cases. The best performance is MLP, followed by GBM and then RF. 

\begin{table}[htbp]
\centering
\caption{RF statistics results for Fit and Unfit class}
\begin{tabular}{l|c|c|c|c}
\hline
\textbf{Class} & \textbf{Sensitivity} & \textbf{Specificity} & \textbf{F1-score} & \textbf{Accuracy} \\ \hline
\hline
\textbf{Fit} & 70.1\% & 94.7\% & 80.5\% & 70.8\% \\ \hline
\textbf{Unfit} & 75.2\% & 28.5\% & 41.3\% & 70.8\% \\ \hline
\end{tabular}
\label{tab_rf_fu}
\end{table}

\begin{table}[htbp]
\centering
\caption{GBM statistics results for Fit and Unfit class}
\begin{tabular}{l|c|c|c|c}
\hline
\textbf{Class} & \textbf{Sensitivity} & \textbf{Specificity} & \textbf{F1-score} & \textbf{Accuracy} \\ \hline
\hline
\textbf{Fit} & 73.1\% & 95.8\% & 82.9\% & 74.0\% \\ \hline
\textbf{Unfit} & 79.8\% & 40.0\% & 45.7\% & 74.0\% \\ \hline
\end{tabular}
\label{tab_gb_fu}
\end{table}

\begin{table}[htbp]
\centering
\caption{MLP neural network statistics results for Fit and Unfit class}
\begin{tabular}{l|c|c|c|c}
\hline
\textbf{Class} & \textbf{Sensitivity} & \textbf{Specificity} & \textbf{F1-score} & \textbf{Accuracy} \\ \hline
\hline
\textbf{Fit} & 75.3\% & 95.4\% & 84.2\% & 75.5\% \\ \hline
\textbf{Unfit} & 77.1\% & 33.1\% & 46.3\% & 75.5\% \\ \hline
\end{tabular}
\label{tab_mlp_fu}
\end{table}

\subsection{Models analyses results}

Note that all the classifiers presented similar results, which indicates that the models are well trained, and the database and the respective separation of it into the subsets was also done correctly.

Although all indicators are essential to study the performance of the models, given the characteristics of the study performed and its applications in real operations, sensitivity and accuracy become relevant indicators. Sensitivity gives us information on the system's ability to detect the classes of interest separately, while accuracy provides an overview of the model, which is relevant in the case of Fit and Unfit class analysis.

For the classification of each condition, RF shows sensitivity results above 70\% for the control and alcohol sets, but the performance declines significantly for drugs and alcohol datasets. However, the misclassifications for the latter two groups are primarily associated with unfit cases. Therefore, if we perform the algorithm analysis for fit and unfit classes, the sensitivity improves substantially, reaching 70.1\% for fit and 75.2\% for unfit. Likewise, the overall accuracy goes to 70.8\%.

The individual results (by class) for GBM show similar behaviour to the ones described for RF. However, the results show an improvement in all groups, outperforming RF results by 3 to 5 points. On the other hand, when analysing the fit and unfit cases, GBM shows 73.1\% and 78.0\% of sensitivity, respectively, and overall accuracy of 74.0\%.

On the other hand, the MLP shows the best results by class for the control group with 75.3\% of sensitivity. For the cases of alcohol and drugs, the sensitivity decreases for the other models, reaching 70.8\% and 29.4\%, respectively. In contrast, the sleep condition remains similar to the other models. The fit and unfit groups analysis obtained a sensibility of 75.3\% and 77.1\%, respectively. The overall accuracy was 75.5\%, the highest of all the models.

\section{Conclusion}
\label{sec:conclusion}

An automatic framework was successfully developed to predict the fitness for duty based on the detection of human fatigue (due to lack of sleep) and the consumption of alcohol and drugs. For practical purposes, the proposed model can estimate whether a given subject is Fit or Unfit. The solution is based on advanced and  state-of-the-art image processing techniques, deep learning algorithms, machine learning, and artificial intelligence, and biometrics.

The behaviour showed essential differences in pupil and iris behaviour under different conditions that may affect the behaviour of the CNS. The differences between the groups were quite remarkable, especially between the control cases and the subjects in the alcohol and drug data subsets. On the other hand, the behaviour of the drowsy or sleep-deprived subjects was closer to the control group. The latter could be due to the reduced number of sleep records in the database.

Based on the obtained results, it can be seen that all models performed similarly for both types of analysis. The best results were shown in the control and alcohol datasets, which is the most extensive and most consolidated subset in our database. At the same time, the drug and sleep records are smaller (17 and 20, respectively, in the test set). This factor reduces the number of records that affect the model's performance, especially considering that we worked with an unbalance database. However, when the analysis was performed only between Fit and Unfit classes, the results improved substantially, and in all cases exceeded 70\%, reaching in some cases values close to 80\%. These numbers represent outstanding results because they indicate that the system is able to detect many individuals who are not fit to perform any activity and, therefore, avoid exposure to any level of risk that could cause harm.

In particular, MLP and GBM showed the best results in all groups, allowing the detection of about 79.8\% and 77.1\%, respectively, of the subjects who could not perform a task safely. This result is an excellent measure and allows us to lower the risk levels of a critical operation that requires fitness in a tangible way.

On the other hand, our results show that it is possible to detect on average 8 out of 10 subjects in Unfit conditions without altering the regular operation of an industry since the system has a specificity of more than 95\% for control (fit) subjects. Furthermore, our database replicates the expected behaviour of this type of operation, in which 10\% to 15\% of the subjects are in Unfit conditions to perform a task. Contrary to previously presented systems, which require immediate and complete attention of the capture subject, our proposed system could be operated as a concurrent observation of the subject, e.g. truck driver or pilot without disturbing his primary duties. Thus over time with the analysis over multiple time window a further improvement of the classification results can be achieved.

The future work will include increasing the database, especially in the cases of subjects who have consumed drugs and subjects sleep-deprived or subjects with daytime sleep due to night shift work. In this way, it is expected to improve the performance of the models in these groups of interest.

On the other hand, the use of deep learning techniques, especially Recurrent Neural Networks (RNN) such as Long Short Term Memory (LSTM)\cite{RAJAMOHANA20212897}, will be explored. This type of network allows working naturally with time series and modelling the temporal behaviour in the internal structure of the network \cite{9016239, ZHANG2021317, ZhongLSTM19}.

\ifCLASSOPTIONcompsoc
  \section*{Acknowledgments}
\else
  \section*{Acknowledgment}
\fi
This work is fully supported by the Agencia Nacional de Investigacion y Desarrollo (ANID) throught FONDEF IDEA N$^{\circ}$ ID19I10118 leading by Juan Tapia Farias - DIMEC-UChile. Further this work has been partially supported by the German Federal Ministry of Education and Research and the Hessian
Ministry of Higher Education, Research, Science and the Arts within their joint support of the National Research Center for Applied Cybersecurity ATHENE.

\ifCLASSOPTIONcaptionsoff
  \newpage
\fi

\bibliographystyle{IEEEtran}
\bibliography{references.bib}
\vspace{-0.4cm}

\begin{IEEEbiography}[{\includegraphics[width=1in,height=1.25in,clip,keepaspectratio]{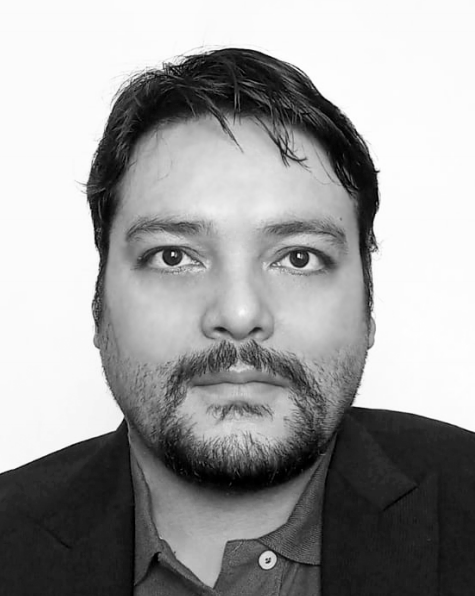}}]%
{Leonardo Causa}
received the P.E. degree in electrical engineering from the Universidad de Chile, in 2012, and the M.S. degree in biomedical
engineering (BME) from the Universidad de Chile, in 2012. He is a Ph.D. (c) in electrical engineering and medical informatics by co-tutelage with the Universidad de Chile and Universite Claude Bernard Lyon 1. His research interests include sleep pattern recognition, signal and image processing, neuro-fuzzy systems applied to classify physiological data, and machine and deep learning. His primary interest is automated sleep-pattern detection and respiratory signal analysis, fitness for duty, human fatigue, drowsiness, alertness, and performance. He is
currently a Researcher with the R\&D Centre TOC Biometrics Company.
\end{IEEEbiography}

\begin{IEEEbiography}[{\includegraphics[width=1in,height=1.25in,clip,keepaspectratio]{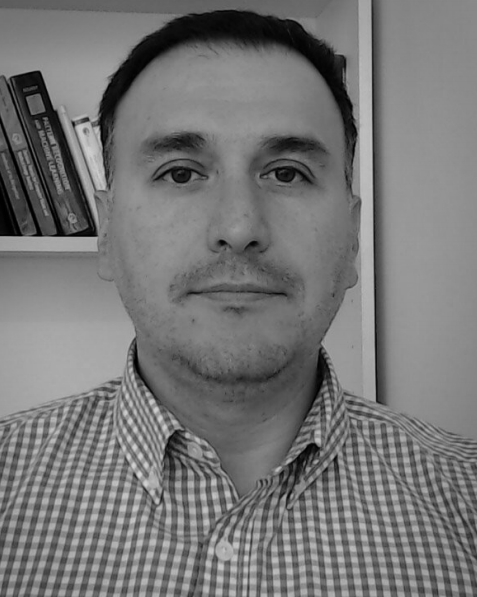}}]%
{Juan Tapia} (Member, IEEE) received a P.E. degree in Electronics Engineering from Universidad Mayor in 2004, a M.Sc. in Electrical Engineering from Universidad de Chile in 2012, and a Ph.D. from the Department of Electrical Engineering, Universidad de Chile in 2016. In addition, he spent one year of internship at University of Notre Dame. In 2016, he received the award for best Ph.D. thesis. From 2016 to 2017, he was an Assistant Professor at Universidad Andres Bello. From 2018 to 2020, he was the R\&D Director for the area of Electricity and Electronics at Universidad Tecnologica de Chile. He is currently a Senior Researcher at Hochschule Darmstadt (HDA), and R\&D Director of TOC Biometrics. His main research interests include pattern recognition and deep learning applied to iris biometrics, morphing, feature fusion, and feature selection. 
\end{IEEEbiography}

\begin{IEEEbiography}[{\includegraphics[width=1in,height=1.25in,clip,keepaspectratio]{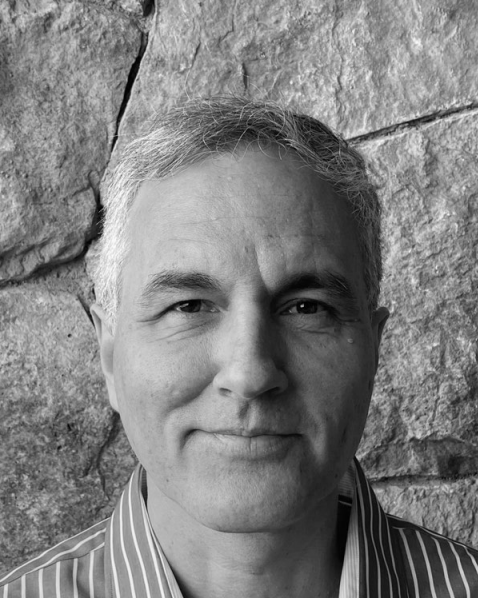}}]%
{Enrique Lopez Droguett}
is currently a Professor with the Civil and Environmental Engineering Department and the Garrick Institute for the Risk Sciences, University of California, Los Angeles (UCLA), USA. He conducts research on Bayesian inference and artificial intelligence supported digital twins and prognostics and health
management based on physics informed deep learning for reliability, risk, and safety assessment of structural and mechanical systems. His most recent focus has been on quantum computing and quantum machine learning for developing solutions for risk and reliability quantification and energy efficiency of complex systems, particularly those involved in renewable energy production. He has led many major studies on these topics for a
broad range of industries, including oil and gas, nuclear energy, defense, civil aviation, mining, renewable, and hydro energy production and distribution networks. He has authored more than 250 papers in archival journals and conference proceedings. He serves in the Board of Directors of the International Association for Probabilistic Safety Assessment and Management (IAPSAM). He is an Associate Editor for the Journal of Risk and Reliability
and the International Journal of Reliability and Safety.
\end{IEEEbiography}
\vspace{-0.5cm}
\begin{IEEEbiography}[{\includegraphics[width=1in,height=1.25in,clip,keepaspectratio]{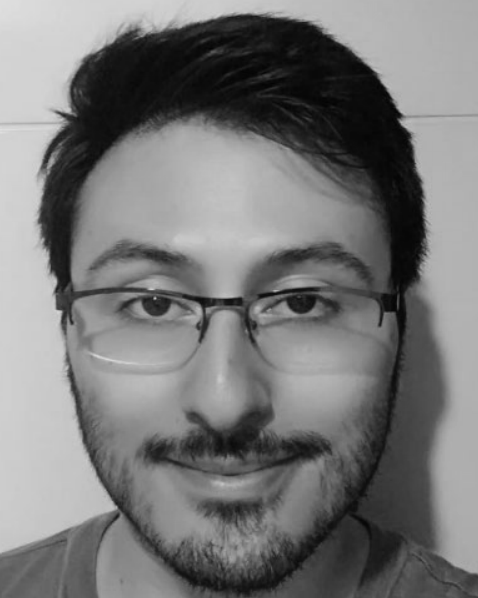}}]%
{Andrés Valenzuela}
received the B.S. degree in computer engineering from the Faculty of Engineering, Universidad Andres Bello, Santiago, Chile, in 2020. He is currently working as a Researcher with the Universidad de Chile and TOC Biometrics, Chile. His main interests include computer vision, pattern recognition, and deep
learning applied to semantic segmentation problems, focusing in NIR and RGB eyes images.
\end{IEEEbiography}
\vspace{-0.5cm}
\begin{IEEEbiography}[{\includegraphics[width=1in,height=1.25in,clip,keepaspectratio]{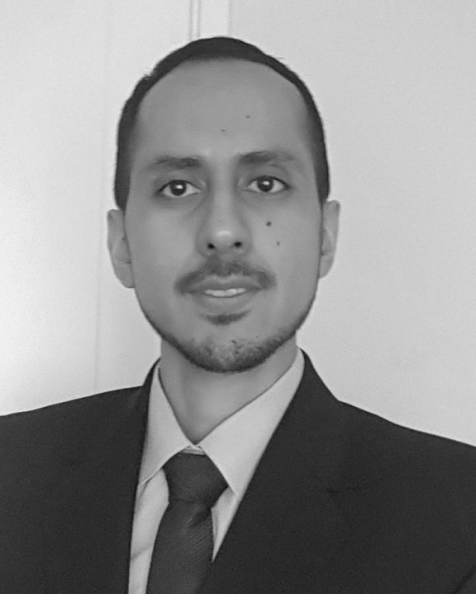}}]%
{Daniel Benalcazar}
(Member, IEEE) was born in Quito, Ecuador, in 1987. He received the B.S. degree in electronics and control engineering from Escuela Politecnica Nacional, Quito, in 2012, the M.S. degree in electrical engineering from The University of Queensland, Brisbane, Australia, in 2014, with a minor in biomedical engineering, and the Ph.D. degree in electrical engineering from
the Universidad de Chile, Santiago, Chile, in 2020. From 2015 to 2016, he worked as a Professor at the Central University of Ecuador. Ever since, he has participated in various research projects in biomedical engineering and biometrics. He is currently working as a Researcher with the Universidad de Chile and TOC Biometrics, Chile.
\end{IEEEbiography}
\vspace{-0.5cm}
\begin{IEEEbiography}[{\includegraphics[width=1in,height=1.25in,clip,keepaspectratio]{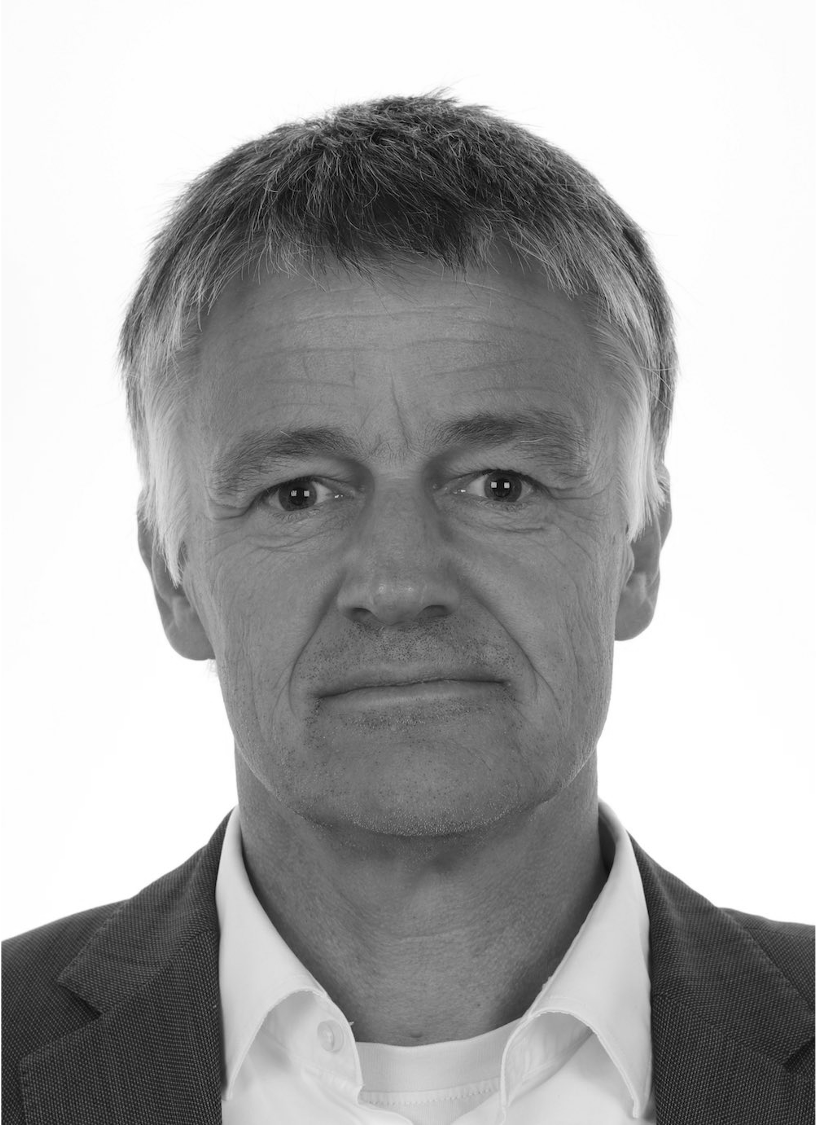}}]%
{Christoph Busch}
(Senior Member, IEEE) is currently a member of the Department of Information Security and Communication Technology (IIK), Norwegian University of Science and Technology (NTNU), Norway. He holds a joint appointment with the Faculty of Computer Science, Hochschule Darmstadt (HDA), Germany. Further, he has been lecturing the course biometric systems at Denmark’s DTU, since 2007. On behalf of the German BSI, he has been the coordinator for the project series BioIS, BioFace, BioFinger, BioKeyS Pilot-DB, KBEinweg, and NFIQ2.0. In the European research program, he was an Initiator of the Integrated Project 3D-Face, FIDELITY, and iMARS. Further, he was/is a Partner in the projects TURBINE, BEST Network, ORIGINS, INGRESS, PIDaaS, SOTAMD, RESPECT, and TReSPAsS. He is also a Principal Investigator with the German National Research Center for Applied Cybersecurity (ATHENE). Moreover, he is a Co-Founder and a member of board of the European Association for Biometrics (www.eab.org) that was established, in 2011, and assembles in the meantime more than 200 institutional members. He coauthored more than 500 technical papers and has been a speaker at international conferences. He is a member of the editorial board of the IET journal.
\end{IEEEbiography}

\end{document}